\documentclass[letterpaper, 10 pt, conference]{ieeeconf}  

\IEEEoverridecommandlockouts                              

\usepackage{graphicx}
\usepackage{amsmath} 
\usepackage{amssymb}  
\usepackage{xfrac}
\usepackage{bbm}
\usepackage[normalem]{ulem}
\usepackage{enumerate}
\usepackage{multirow}
\usepackage{tabularx}
\usepackage[table]{xcolor}
\usepackage{array} 
\usepackage{makecell}
\usepackage{longtable}
\usepackage{booktabs}
\usepackage{url}
\usepackage{hyperref}

\newcolumntype{L}[1]{>{\raggedright\arraybackslash}p{#1}}
 
\definecolor{successgreen}{rgb}{0.9, 1.0, 0.9}
\newcommand{\bt}{\textcolor{blue}}

\title{\LARGE \bf
Whole-Body Planning for Humanoids Navigating Confined Spaces via 
Self-Collision Avoidance References
}

\author{Carlos Gonzalez$^{1}$ and Luis Sentis$^{1}$%
\thanks{$^{1}$Authors are with the Department of Aerospace Engineering and Engineering
Mechanics, The University of Texas at Austin, TX 78712, USA.
        {\tt\small \{carlos.gonzalez,lsentis\}@utexas.edu}}%
}

\begin{document}

\maketitle
\footnotetext[2]{Supplementary videos are available at: 
\href{https://carlosiglezb.github.io/confined-space-wbp-humanoid/}{https://carlosiglezb.github.io/confined-space-wbp-humanoid/} \\
\thanks{This work has been submitted to the IEEE for possible publication. Copyright may be transferred without notice, after which this version may no longer be accessible.}
}

\thispagestyle{empty}
\pagestyle{empty}

\begin{abstract}
Humanoid locomotion in highly confined environments requires navigating dense environmental obstacles and complex self-collision bounds while maintaining multi-contact dynamic feasibility. Traditional trajectory optimizers frequently struggle in these restricted spaces, as navigating the large collision space with splines on particle abstractions is insufficient and leads to poor local minima. To address this, we propose a three-stage whole-body planning framework that formulates kinematic path planning directly over kinematically reachable rigid-body volumes. By integrating differentiable collision avoidance into a reachability-constrained formulation, our framework synthesizes volume-informed guides that reliably guide a full-order trajectory optimizer over long horizons. We show that these optimized plans serve as high-quality references to train a residual reinforcement learning policy for robust online execution. We validate our approach on the Unitree G1 humanoid across three benchmark testbeds exceeding NIST emergency response standards, achieving restricted confinement ratios ($C_r < 1.5$). Our framework generates feasible trajectories across 12-to-18-second tasks with complex foot and hand contacts where standard baselines fail, while the learned policy successfully tracks these plans under extensive domain randomization in physics simulation.
\end{abstract}
\section{Introduction}

Legged locomotion on structured terrains 
has matured rapidly in recent years, with fluid behaviors ranging from human-like walking on flat
ground~\cite{ames_human-inspired_2014} to agile locomotion over large obstacles~\cite{Wu2026PerceptiveMatching}.
Learning-based approaches have streamlined the execution of dynamic motions that were 
traditionally difficult to realize with classical planning schemes, 
at the expense of relying on high-quality motion data.
At the same time, advances in convex optimization and differentiable geometry for motion planning 
have reduced solve times and minimized manual parameter tuning when generating 
collision-free trajectories. These developments provide a strong foundation for 
synthesizing kinematically and dynamically consistent motions in highly constrained 
environments, such as confined spaces.

Navigating confined spaces presents severe challenges for humanoid robots: 
the collision-free configuration space degrades into narrow, non-convex manifolds 
where slight posture deviations cause unwanted environmental collisions that risk 
destabilizing the robot. 
Furthermore, long-horizon planning is essential, as myopic or short-horizon schemes 
fail to foresee when current footholds or upper-body postures lead to kinematic or dynamic dead-ends. 
The scarcity of high-quality demonstration data in confined settings limits the 
direct applicability of data-driven and imitation-based 
frameworks~\cite{Yang2025OmniRetarget:Interaction,wang_hil_2026}, which require precise 
motion retargeting that strictly respects the robot's kinematic reachability and torque limits. 
Conversely, classical trajectory optimization formulations often struggle with dense obstacle scenes: 
they typically neglect detailed collisions to maintain reasonable compute 
times~\cite{Winkler2018GaitParameterization,Wang2020Multi-FidelityLocomotion}, 
rely on overly simplified bounding volumes~\cite{taouil_physically_2025}, 
or require sophisticated initializations to converge to feasible solutions~\cite{Dai2014}.


\begin{figure}[t]
\centering
\includegraphics[width=0.9\linewidth]{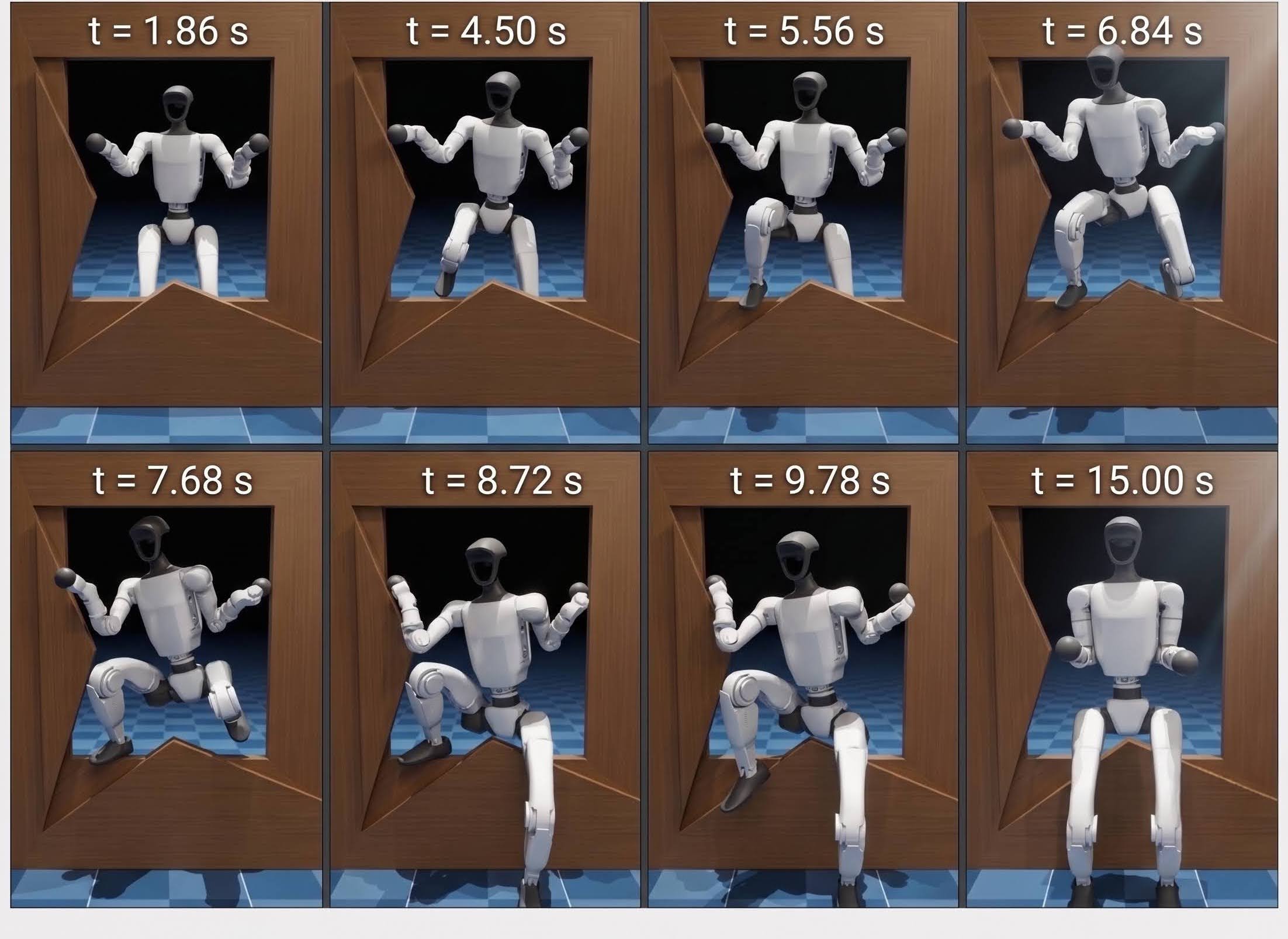}
\caption{\textbf{Whole-body motion through an obstructed hole.} 
The G1 humanoid navigates a non-convex aperture in a confined space. Our proposed WBP
generates a kinematically and dynamically feasible plan that the robot is able to track online
despite the terrain complexities: it balances balances on non-coplanar hand and foot contacts 
to cross the opening ($t=4.50\text{--}7.68\,\text{s}$), then lands in a 
kinematically constrained configuration and resolves into a squared-up 
stance ($t=8.72\text{--}15.00\,\text{s}$).}
\label{fig:teaser}
\end{figure}

To address these challenges, we introduce a morphology-aware Whole-Body Planning (WBP)
and Control framework for humanoid robots in tight, geometrically complex environments, 
such as those defined by NIST emergency response
standards~\cite{2020StandardStairs/Landings,2021TestObstacles}. 
At the core of our approach is a novel kinematic seeding mechanism that lifts 
particle-based convex relaxations~\cite{Gonzalez2024GuidingOptimization}
to full 3D rigid-body representations. By embedding differentiable collision 
detection~\cite{Tracy2023DifferentiablePrimitives}
directly into this relaxed geometric search, our planner effectively traverses 
narrow non-convex configuration manifolds to discover topologically distinct, 
kinematically feasible posture seeds, automatically uncovering diverse 
locomotion strategies without requiring manual initial guesses.
These kinematic seeds are subsequently transcribed into dynamically consistent 
whole-body trajectories that push the robot to its physical limits.

Our main contributions are summarized as follows:

\begin{enumerate}[1)]
    \item \textbf{Geometry-Aware Kinematic Reference Generation.} We efficiently extend 
    the constrained particle planning strategy in~\cite{Gonzalez2024GuidingOptimization} to
    constrained rigid bodies.
    Our approach promotes the discovery of kinematically feasible seeds
    for diverse locomotion styles to guide Self-Collision Avoidance (SCA) motions
    in confined spaces.
    \item \textbf{Dynamically Consistent Confined-Space Planning.} 
    A morphology-aware optimization pipeline that transcribes complex kinematic collision-free 
    references into dynamically consistent whole-body trajectories for humanoids.
    \item \textbf{Closed-Loop Whole-Body Tracking Policy.}
    A fast-converging learned policy that tracks the generated WBPs 
    in simulation under sensor noise, modeling uncertainties, and dynamics randomization.
    \item \textbf{Long-Horizon Benchmark in Standardized Confined Spaces.} 
    We validate our framework on the Unitree G1 humanoid, generating stable, long-horizon
    plans ($12$--$18\,\text{s}$) across three confined environments that meet and exceed 
    NIST emergency response standards~\cite{2020StandardStairs/Landings, 2021TestObstacles}, 
    where baseline spline-guided planners fail to converge.
\end{enumerate}

\begin{figure}
    \centering
    \includegraphics[width=0.95\linewidth]{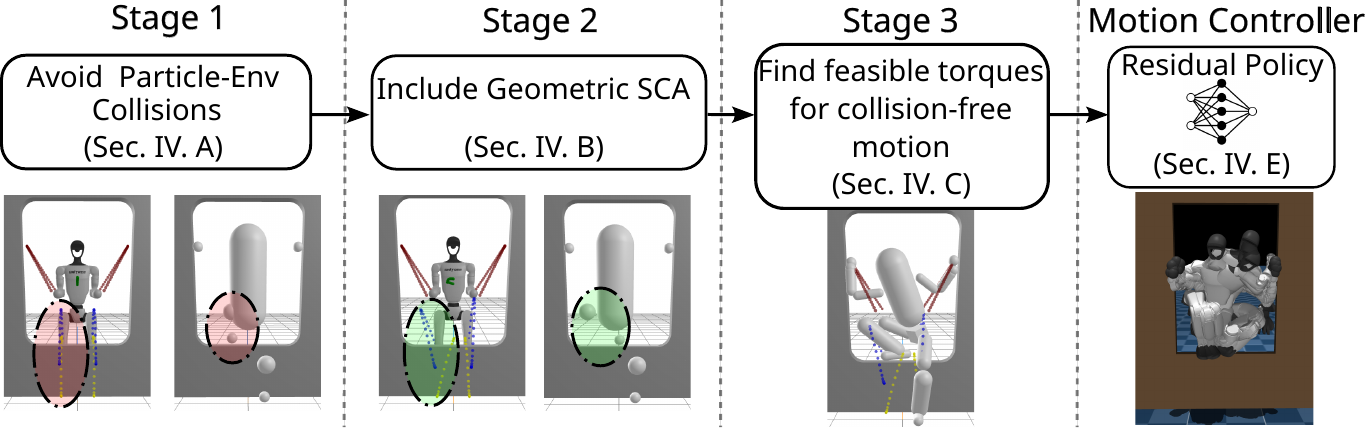}
    \caption{\textbf{Sequential planning pipeline.} Stage 1: Environment-aware TO
    for planning frames. Stage 2: SCA refinement using primitive proxy geometries. 
    The right leg trajectories are notably modified by this stage. 
    Stage 3: Dynamically feasible WBP with more articulated collision model. 
    Motion Controller: Residual policy tracking the computed plan.}
    \label{fig:architecture}
\end{figure}

\section{Related Work} \label{sec:related_work}

Finding a feasible motion that moves a robot from an initial to a goal state by making and breaking contacts with the environment is known as the multi-contact planning problem. Existing approaches to solve this problem span optimization-based formulations, imitation learning from demonstrations, and end-to-end reinforcement learning.

\subsection{Optimization-Based Motion Planning} \label{subsec:rel_optimization}
Multi-contact motion generation typically follows a stance-before-motion paradigm~\cite{Wensing2023Optimization-BasedRobots, Bretl2006MotionProblem}. While dedicated contact planners search for contact sequences using reachability graphs~\cite{Tonneau2018AnRobots}, mixed-integer optimization~\cite{Deits2014FootstepOptimization}, or implicit sampling~\cite{Howell2022PredictiveMuJoCo}, this work focuses specifically on WBP given candidate sequences. Our goal is to address the remaining continuous non-convex challenge: discovering feasible joint and torque trajectories through narrow passages.

State-of-the-art WBP relies on Trajectory Optimization (TO) operating over either 
reduced-order models~\cite{Winkler2018GaitParameterization,Wang2020Multi-FidelityLocomotion} 
or full-body formulations~\cite{Gonzalez2024GuidingOptimization}. Reduced Order Models (ROMs) are computationally efficient but neglect link collisions, making them unsuitable for confined 
environments. Conversely, full-body TO captures a more representative geometry but 
struggles with collision enforcement in tight spaces. Existing methods handle collisions either 
via soft penalty barriers~\cite{Chiu2022AManipulation}, which allow residual penetrations, 
or hard inequality constraints~\cite{Schulman2014MotionChecking}, which trap gradient-based 
solvers in local minima without high-quality initializations.

\subsection{Imitation-Based Learning} \label{subsec:rel_imitation}
To overcome the challenges in generating feasible multi-contact trajectories,
data-driven and imitation-based learning methods retarget human motion capture data or 
imitate pre-generated trajectories to synthesize whole-body 
behaviors~\cite{Yang2025OmniRetarget:Interaction, Peng2018DeepMimic:Skills}. 
Recent frameworks use trajectory-tracking imitation~\cite{Liu2025Opt2Skill:Loco-Manipulation} 
to transfer complex whole-body skills to physical humanoids.

While imitation learning excels at highly athletic 
maneuvers~\cite{wang_hil_2026,Wu2026PerceptiveMatching}, it heavily relies 
on the availability of rich, physically feasible reference datasets.
Acquiring this data in confined spaces is extremely difficult due to unnatural joint configurations 
needed to traverse them, in addition to potential kinematic constraint mismatches. 

\subsection{End-to-End Deep Reinforcement Learning} \label{subsec:rel_rl}
Deep Reinforcement Learning (RL) has demonstrated impressive results in dynamic bipedal locomotion 
and perceptive navigation~\cite{Hoeller2024ANYmalRobots, Zhuang2025HumanoidLearning, Kumagai2024ReinforcementHumanoids}. 
End-to-end policies directly map onboard sensory observations to motor commands, enabling 
real-time adaptation over uneven terrain and 3D 
obstacles~\cite{Xu2024DexterousLearning, Buchanan2021PerceptiveSpaces, Miki2024LearningRepresentation}.

Despite these advances, navigating complex and tight spaces requiring 
self- and environment-collision awareness purely through reward exploration leads to extreme 
sample inefficiency, local minima, and dynamic instability.

This work bridges these gaps by efficiently synthesizing morphology-aware guides that a TO 
can leverage to lead the robot through the confined space. 
Our pipeline generates plans directly over kinematically reachable rigid-body volumes using 
differentiable collision detection~\cite{Tracy2023DifferentiablePrimitives}. 
This approach provides the non-myopic seed required to generate high-fidelity reference trajectories
that subsequently enables effective RL policy training for online execution.

\section{Problem Statement}\label{sec:problem}
In multi-contact legged locomotion, the overall planning problem can be stated as
follows:
\begin{subequations}
    \begin{align}
        \underset{\substack{x(\cdot), u(\cdot), \\
        t_1, \cdots, t_{n_{\mathrm{ph}}}, \\
        n_{\mathrm{ph}}}}{\mathrm{minimize}}       & \quad \sum\limits_{i=1}^{n_{\mathrm{ph}}} \int\limits_{t_{i-1}}^{t_i} 
                                        \ell_i\left( x(t), u(t) \right) \mathrm{d}t + \Phi_{i} (x(t_i)) \label{subeq:mcp-dyn} \\
        \mathrm{subject~to}     & \quad \dot{x}(t) = f_i(x(t), u(t)) \\
        & \quad g_i(x(t), u(t)) = 0         \label{subeq:mcp-eq-constraint}\\
        & \quad h_i(x(t), u(t)) \leq 0      \label{subeq:mcp-ineq-constraint}\\
        & \quad x(t_i^+) = \rho_i(x(t_i^-))    \label{subeq:mcp-reset-map}
    \end{align}
    \label{eq:mcp-ct}
\end{subequations}
where the optimization variables are the state, $x(t)$, the control input, $u(t)$,
the contact phase transition times $t_i, ~\forall~i = 0, \cdots, n_{\mathrm{ph}}$, and
the total number of contact phases, $n_{\mathrm{ph}}$.
The running and terminal costs are $\ell_i(\cdot, \cdot)$ and $\Phi_i(\cdot)$,
respectively. The dynamics of the system are dictated according 
to its current contact phase~\eqref{subeq:mcp-dyn}, where a specific set of 
equality~\eqref{subeq:mcp-eq-constraint} and 
inequality~\eqref{subeq:mcp-ineq-constraint} conditions may apply. 
When a change in contacts occurs, i.e., at
$t_1, \ldots, t_{n_{\mathrm{ph}}-1}$,
a new state, $x(t^+_i)$, is obtained through some reset map (e.g., impulse dynamics model),
as shown in \eqref{subeq:mcp-reset-map}.

Problem~\eqref{eq:mcp-ct} is computationally intractable due to the 
high-dimensional continuous state space, the combinatorial complexity of the 
discrete contact sequences, and the non-convexity of environmental collision 
constraints. We address this by partitioning~\eqref{eq:mcp-ct} into two 
sub-problems: the contact sequence selection and the whole-body motion synthesis. 
In this work, we focus on the latter since the former
can be effectively explored using existing methods such as tree
search methods~\cite{Taouil2026MotionDisco:Loco-Manipulation}.
Hence, we assume a provided 
sequence of $n_{\mathrm{ph}}$ contact phases $\{ c_i \}_{i=1}^{n_{\mathrm{ph}}}$
with durations $T_i:=t_i - t_{i-1}$, where each $c_i$ represents 
the $i^{\mathrm{th}}$ contact phase from the set of possible contact
combinations $\mathcal{C}$.
This modular approach allows the planner to remain agnostic to the contact 
sequence generator, provided these are kinematically reachable.
With these assumptions, we discretize problem~\eqref{eq:mcp-ct} and solve the following
simplified whole-body planning problem:
\begin{subequations}
    \begin{align}
        \underset{x, u}{\mathrm{minimize}}       & \quad \sum\limits_{i=1}^{n_{\mathrm{ph}}} \sum\limits_{k=N_{i-1}}^{N_i - 1}  \ell_k\left( x_k, u_k \right) + \Phi_{N_{n_{\mathrm{ph}}}} (x_{N_{n_{\mathrm{ph}}}}) \\
        \mathrm{s.t.}     & \quad x_{k+1} = f^d_i(x_k, u_k), \quad \forall\, k \in [N_0, N_{n_{\mathrm{ph}}}] \label{subeq:wbp-dyn} \\
        & \quad x_k \in \mathcal{X}_{\lim}, \quad u_k \in \mathcal{U}, \quad \lambda_k \in \mathcal{K}_i \label{subeq:wbp-ineqs}\\
        & \quad x_k \in \mathcal{X}_{\mathrm{cfree}} \cap \mathcal{X}_{\mathrm{sca}} \label{subeq:wbp-cfree}
    \end{align}
    \label{eq:wbp-dt}
\end{subequations}

%
%
Contact phase $i$ consists of $N_i - N_{i-1}$ knots, where $N_i \in \mathbb{Z}^+$
is the time index at which contact phase $i$ ends.
The discrete dynamics~\eqref{subeq:wbp-dyn} dictates the evolution of
the state in each contact phase $i$.
Joint positions and velocities are constrained
to be within their respective limits in $\mathcal{X}_{\lim}$. 
Similarly, torques are constrained to be
within their limits, $\mathcal{U}$.
The contact forces must remain within their respective friction cones, $\mathcal{K}_i$.
Constraint~\eqref{subeq:wbp-cfree} restricts the robot to configurations that 
are free of self-collision, $\mathcal{X}_{\mathrm{sca}}$, and environmental 
interference, $\mathcal{X}_{\mathrm{cfree}}$, with the exception of the 
specific contact patches scheduled for navigation.
We highlight here that the sets
$\mathcal{X}_{\mathrm{cfree}}$ and $\mathcal{X}_{\mathrm{sca}}$ become
more restrictive when navigating confined spaces and are
in general non-convex, hence our aim to guide the solver away from
local minima induced by these constraints.
We omit impulse dynamics under the assumption of 
negligible impact velocities during the planned contact transitions.

The plan obtained from solving~\eqref{eq:wbp-dt} to navigate
a particular environment must be tracked online
to account and correct for model discrepancies in real-time.
For this purpose, we seek a controller that outputs action 
$a \sim \pi(a_t | s_t)$ given the current state $s_t \in \mathcal{S}$. 
In practice, this policy is conditioned on noisy observations 
$o_t \in \mathcal{O}$, leading to a policy of the form $\pi (a_t | o_t)$.

\section{Approach}\label{sec:approach}
The proposed framework to solve~\eqref{eq:wbp-dt} is shown in
Fig.~\ref{fig:architecture}. The planning problem is broken down into three
stages. Stage 1 (Sec.~\ref{subseq:mfpp}), 
introduced in~\cite{Gonzalez2024GuidingOptimization},
generates smooth, environment-collision-free paths for the task-space frames,
e.g., torso, feet, palms, while ensuring kinematic reachability.
Stage 2 (Sec.~\ref{subseq:sca}) introduces our refinement
process to incorporate proxy collision geometries to steer these
trajectories away from self-collisions while allowing for desired
stylized behaviors.  
Stage 3 (Sec.~\ref{subseq:dyn-to}) uses these refined paths as guides
in a TO problem to find feasible torque
profiles consistent with the full robot dynamics~\eqref{subeq:wbp-dyn}
and that respect the physical limits~\eqref{subeq:mcp-ineq-constraint} and physical
interaction with its environment~\eqref{subeq:wbp-cfree}.
The pipeline integration is
detailed in Sec.~\ref{subsec:pipeline}.
Lastly, these plans are used to learn an online policy that follows
the desired motion online under potential real-world perturbations
(Sec.~\ref{subsec:training}).


\subsection{Kinematic Planning Avoiding Environment Collisions~\cite{Gonzalez2024GuidingOptimization}}
\label{subseq:mfpp}

Stage 1 generates smooth, collision-free task-space reference paths for the torso, feet, and palms while enforcing kinematic reachability. The environment is represented as primitive convex volumes (e.g., IRIS regions~\cite{deits15_iris}) assigned to each planning frame. A search-based planner determines the region traversal sequence, allocating phase durations $T_i$ proportional to path lengths within each region.

Task-space trajectories are parameterized as B\'ezier curves with control points $\beta$. The optimal control points are obtained by solving a Second-Order Cone Program (SOCP):
\begin{equation}
    \underset{\beta}{\mathrm{minimize}} \quad \text{path\_len\_dynamics}(\beta; w_1) + \mathrm{RigL}(\beta; w_2)
    \label{eq:kin_bezier}
\end{equation}
subject to contact schedule, reachability, and collision-free bounds within the IRIS regions. Here, $\mathrm{RigL}(\beta; w_2)$ is a logarithmic penalty on knee-to-ankle distance variance, forming a convex relaxation of the rigid link constraint. Non-contacting environment surfaces are padded by link radii, yielding a smooth kinematic reference $x^{\mathrm{des}}(t)$.

\subsection{Differentiable Self-Collision Avoidance} \label{subseq:sca}
\begin{figure}
    \centering
    \includegraphics[width=0.5\linewidth]{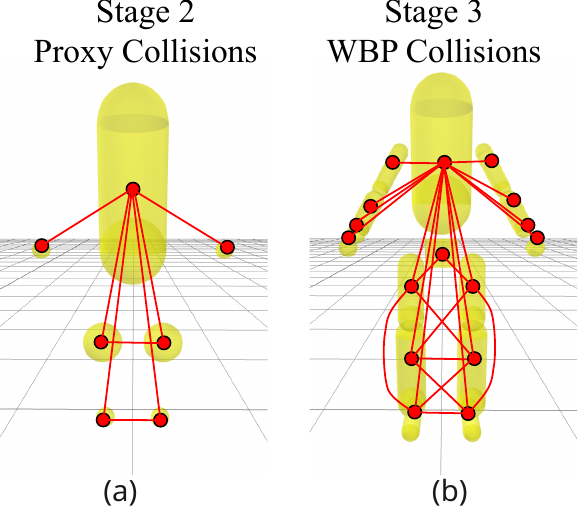}
    \caption{\textbf{Collision models used in planning pipeline.} 
    (a) Stage 2: Primitive-based (capsules and spheres) geometries for kinematic guides. 
    (b) Stage 3: Refined joint-level bodies (capsules, spheres, and a pelvis box). 
    Red lines denote active collision pairs.}
    \label{fig:collision-models}
\end{figure}
%
%
This module efficiently adjusts the kinematic paths from Sec.~\ref{subseq:mfpp} to 
account for the volumes of the rigid bodies moving along them. Specifically, 
these bodies must satisfy SCA while preserving their reachability properties. 
Building upon the differentiable collision framework in~\cite{Tracy2023DifferentiablePrimitives}, 
the minimum distance between these primitive pairs is computed by solving 
the following SOCP:
\begin{equation}
    \begin{aligned}
        \underset{x, \alpha}{\mathrm{minimize}}       & \quad  \alpha \\
        \mathrm{subject~to}     & \quad x \in \mathcal{S}_1 (\alpha) \\
                                & \quad x \in \mathcal{S}_2 (\alpha) \\
                                & \quad \alpha \geq 0
    \end{aligned}
    \label{eq:collision-detection}
\end{equation}
where $x \in \mathbb{R}^3$, and
$\mathcal{S}_1$ and $\mathcal{S}_2$ are convex shapes expanded by a factor of $\alpha$.
If $\alpha^* >1$, it means that the bodies $\mathcal{S}_1$ and $\mathcal{S}_2$
must be expanded for them to intersect, hence they are not in collision.
Alternatively, if $\alpha^* \leq 1$, it means that the bodies are already
in collision. 

The benefit of writing collision detection as~\eqref{eq:collision-detection} is that
its gradient w.r.t. position is readily obtained through
the Lagrangian and its primal-dual solution.
This means that, instead of padding the robot with several spheres to evaluate
collisions~\cite{Buchanan2021PerceptiveSpaces}, 
we can use a single rigid body primitive and perform
the required computations on a single body pair.
At the same time, instead of numerically estimating the derivatives by 
finite difference on perturbations at a nominal state, we can simply
compute the gradients directly.

As illustrated in Fig.~\ref{fig:collision-models}, we approximate the humanoid’s 
torso with a capsule and its palms, knees, and ankles with spheres. 
The capsule and sphere primitives are expressed by
\begin{align}
    \| x - ( r_1 + \gamma \hat{b}_z ) \|_2 &\leq \alpha R, &
    - \alpha \frac{L}{2} \leq \gamma &\leq \alpha \frac{L}{2} \\
    \| U (x-r_2) \|_2 &\leq \alpha, &
\end{align}
respectively, where $\hat{b}_z = Q \left[0, 0, 1 \right]^\top$, 
$Q \in \mathbb{R}^{3 \times 3}$ corresponds to
the rotation of the capsule w.r.t. world frame,
$\gamma \in \mathbb{R}$ is an
additional slack variable, $r_1, r_2 \in \mathbb{R}^3$ are the coordinates of the origin
of the capsule and sphere, respectively, 
$R$ and $L$ are the radius and cylinder length of the capsule, 
and
$U \in \mathbb{R}^{3 \times 3}$ is the Cholesky factorization of the sphere.
These are our corresponding set constraints in~\eqref{eq:collision-detection}.

We include the rigid body collision detection~\eqref{eq:collision-detection} 
as a constraint in problem~\eqref{eq:kin_bezier}, leading to a constraint 
of the form $\alpha(\beta) \geq 1$ 
for all control points in~\eqref{eq:kin_bezier}. Note that this
constraint is parameterized by the origin of the primitive shape, 
which corresponds to our control points, $\beta$.

Since we are performing gradient-based optimization, we must also provide
the solver with the respective gradient of this new constraint w.r.t.
the optimization variable $\beta$.
This is conveniently obtained 
by
invoking the Implicit Function Theorem (IFT), using the KKT condition
on the cone constraint
$(h - Gx^\star) \circ z^\star=0$ for the corresponding collision pairs
as the implicit functions, where $(x^{\star}, z^{\star})$
are the corresponding primal-dual solution. 
The gradients $\frac{\partial \alpha}{\partial \beta}$ 
are then computed directly from the IFT on parameter $\beta$.
Applying these constraints at the discrete B\'ezier control points $\beta$ 
does not strictly guarantee collision avoidance for the continuous path. 
However, the convex hull property of B\'ezier curves ensures that the 
resulting trajectory remains closely bounded by these points. 
In the constrained environments considered here, this approximation 
provides a computationally efficient proxy that effectively guides the 
motion toward a feasible basin for the high-fidelity refinement in Stage 3.

In order to solve problem~\eqref{eq:kin_bezier} augmented with SCA,
we formulate the problem using CasADi, which allows us to perform
Automatic Differentiation on the constraints already included
therein. The SCA constraints are constructed through
a custom callback that solves~\eqref{eq:collision-detection} within
the evaluate call. This stores the corresponding primal-dual variables,
which are then used in the call to the gradient of the function. Moreover,
in order to make full use of our sequential framework, we employ
the warm-starting routine using both primal-dual solution 
from~\eqref{eq:kin_bezier} solved without the SCA, and heuristically
zero-pad the dual variables of the SCA constraint.
This SCA constraint is applied between the collision pairs 
indicated in Fig.~\ref{fig:collision-models}(a).


\subsection{Full-Order Dynamics Trajectory Optimization} \label{subseq:dyn-to}
Once the paths of the planning frames have been adjusted so that they
are all kinematically reachable, consistent with the contact
plan, and avoid self- and environment collisions, we
compute a dynamically feasible plan guided by these trajectories.
This amounts to solving problem~\eqref{eq:wbp-dt}, which we aim
to speed-up by guiding the solver's landscape through the
near-feasible paths resulting from Sec.~\ref{subseq:sca}.

We use the whole-body dynamics model of the humanoid, with state vector
$x = [q^\top ~v^\top]^\top$. The generalized configuration is 
$q \in \mathrm{SE}(3) \times \mathbb{R}^{n_j}$ representing the
floating base and the $n_j$ joint, while the generalized velocity is
$v \in \mathbb{R}^{n_v}$, where $n_v=6 + n_j$.
The control input $u \in \mathbb{R}^{n_u}$ corresponds to the actuated
joint torques.

\subsubsection{Costs} \label{subseq:dyn-to-costs}
To guide the whole-body trajectory optimizer along the Stage 2 references, we formulate 
running cost $\ell_k$ and terminal cost $\Phi_N$ as:
\begin{equation}
\begin{aligned}
    \ell_k(x_k, u_k) &= \sum_{f \in \mathcal{F}} w^{\mathrm{goal}} \| \mathrm{FK}(x_k) - x^{f,\mathrm{des}}_k \|_{W_f}^2 \\
    &\quad + w^{\mathrm{xReg}} \| x_k - x^{\mathrm{ref}}_k \|_{W_x}^2 + w^{\mathrm{uReg}} \| u_k \|_{W_u}^2, \\
    \Phi_N(x_N) &= \sum_{f \in \mathcal{F}} w^{\mathrm{goal}}_N \| \mathrm{FK}(x_N) - x^{f,\mathrm{des}}_N  \|_{W_f}^2 \\
    &\quad + w^{\mathrm{xReg}}_N \| x_N - x^{\mathrm{ref}}_N \|_{W_x}^2.
\end{aligned} 
\label{eq:wbp-costs}
\end{equation}
where $\mathrm{FK}(\cdot)$ evaluates task-space forward kinematics, 
$x^{f,\mathrm{des}}_k$ is the desired positions of task frame $f \in \mathcal{F}$
obtained from its corresponding B\'ezier curves,
and $W_f, W_x, W_u$ are positive definite weighting matrices.

%
%
%
where $w^{\mathrm{goal}}, w^{\mathrm{goal}}_N \in \mathbb{R}$ and $W_f \in \mathbb{R}^{6 \times 6}$ are weights,
$\mathrm{FK} : \mathbb{R}^{7 + n_j} \mapsto \mathbb{R}^3 \times \mathfrak{so}(3)$ 
computes the forward kinematics
based on the current state configuration,
$x^{f,\mathrm{des}}_{k} \in \mathbb{R}^3 \times \mathfrak{so}(3)$ is constructed
by combining the
B\'ezier curves for all frames in $\mathcal{F}$, 
as computed in Sec.~\ref{subseq:sca}, and the orientations of the
contact surfaces from the environment.
Since the B\'ezier curves 
can be evaluated
at any point in time $t \in [t_0, t_{n_{\mathrm{ph}}}]$,
it provides the solver with smooth translation references.

%
%
%
%

\subsubsection{Constraints}

In order to search over physically possible motions, we apply joint
position and velocity limits in the form of box constraints, $\mathcal{X}_{\lim}$,
for each joint, excluding the free floating base. 
Similarly, $\mathcal{U}$ correspond to box constraints for the joint torques. 

Friction is enforced in two different forms: as a contact
wrench cone constraint for the feet contact patches and as 
a friction cone constraint for the hand
point contacts. In both cases, the contact
surfaces used as references are the one obtained from the
segmentation layer, required as inputs to our planning
algorithm in Sec.~\ref{subseq:mfpp}.

In order to strictly account for environment and self-collision avoidance,
in addition to the guides created
in Sec.~\ref{subseq:sca}, we use a refined collision representation 
of the robot (see Fig.~\ref{fig:collision-models}(b)) for strict 
collision avoidance. This is implemented as a hard constraint 
in the refined WBP as in~\cite{Haffemayer2024ModelArm}
using the signed distance $d(A, B) = \sigma \| w_A - w_B \|$, 
where $\sigma = -1$ when in collision and $\sigma = 1$ otherwise,
and computing its derivative w.r.t. the robot configuration via
\begin{equation}
    \frac{\partial d}{\partial q} = \frac{\sigma}{d} \left( w_B - w_A\right) 
        \left( \frac{\partial w_B}{\partial q} - \frac{\partial w_A}{\partial q}\right) 
\label{eq:joint-level-sca}
\end{equation}
where $\frac{\partial w_A}{\partial q}$ is the Jacobian of the (closest) witness 
point attached to body A.

We use the Constrained SQP solver~\cite{Jordana2025Structure-ExploitingControl}
to solve~\eqref{eq:wbp-dt}. 
The torque, environment, and SCA constraints are enforced as hard constraints while
the friction constraints are implemented as soft constraints with a Barrier.

\subsection{Pipeline for Fully SCA Motions} \label{subsec:pipeline}
To ensure robust convergence of the high-fidelity problem~\eqref{eq:wbp-dt}, 
we implement a two-pass optimization strategy. In the first pass, 
we solve for dynamically consistent joint torques and contact 
forces while tracking the guides from Stage 2 via $x^{f,\mathrm{des}}_k$
in~\eqref{eq:wbp-costs} while
neglecting hard collision constraints.
This allows the solver to prioritize satisfying the nonlinear 
dynamics~\eqref{subeq:wbp-dyn} and respecting state and torque 
limits~$\mathcal{X}_{\lim}$ and $\mathcal{U}$, respectively, while remaining in the 
proximity of the collision-free space. In the second pass, 
the resulting trajectory serves as a warm-start for a final optimization 
where collision avoidance is enforced as hard inequality 
constraints~\eqref{eq:joint-level-sca} using the SCA model in Fig.~\ref{fig:collision-models}. 
This sequential approach prevents numerical 
instability and avoids falling in an infeasible start trap since the solver 
begins the final pass within a basin of attraction that is 
already both dynamically plausible and nearly collision-free.

To initialize Stage 3, target end-effector poses from the contact sequence are solved 
via Inverse Kinematics to yield nominal joint configurations with orientation preferences
along the respective contact surfaces. 
Corresponding quasi-static torques are then computed under contact friction constraints, 
providing a physically grounded seed for the solver.

\begin{table}[t]
\caption{Observation Noise Terms and Domain Randomization}
\label{tab:dr-terms}
\centering 
\small
\setlength{\tabcolsep}{4pt}
\renewcommand{\arraystretch}{1.0}
\begin{tabular}{@{}lll@{}}
  \toprule
  \textbf{Parameter} & \textbf{Type} & \textbf{Range / Std.} \\
  \midrule
  Base Vel. (Lin. / Ang.) & Additive & $\pm 0.15$\,m/s, $\pm 0.15$\,rad/s \\
  Projected Gravity & Additive & $\pm 0.05$ \\
  Joint Pos. / Vel. & Additive & $\pm 0.1$\,rad, $\pm 0.1$\,rad/s \\
  \midrule
  Link Mass \& Inertia & Scaling & $[0.85, 1.15]$ \\
  Tangential Friction & Absolute & $[0.6, 0.9]$ \\
  Base Push Vel. $(x,y)$ & Additive & $(\pm0.1, \pm0.4)$\,m/s \\ 
  Hand Push Force & Additive & $[15, 30]$\,N ($0.2\text{--}0.5$\,s pulse) \\
  Spawn Pos. $(x,y)$ & Absolute & $X_{\mathrm{dom}} \times Y_{\mathrm{dom}}$ \\
  \bottomrule
\end{tabular}
\end{table}

\begin{table}[t]
\caption{Reward Terms for the G1 Residual Tracking Policy}
\label{tab:residual-rewards}
\centering 
\small
\setlength{\tabcolsep}{2.5pt}
\renewcommand{\arraystretch}{1.0}
\begin{tabularx}{\linewidth}{@{} X l r @{}}
  \toprule
  \textbf{Term} & \textbf{Kernel Type} & \textbf{Weight} \\
  \midrule
  Joint Position & Gaussian ($\sigma = 0.3$\,rad) & $1.0$ \\
  Center-of-Mass Pos. & Gaussian ($\sigma = 0.05$\,m) & $1.5$ \\
  End-Effector Pos. & Gaussian ($\sigma = 0.05$\,m) & $2.0$ \\
  Contact Force (Foot/Hand) & Gaussian ($\sigma = 60/20$\,N) & $1.0$ \\
  \midrule
  Friction-Cone Violation & Squared-Hinge & $-0.5$ \\
  Sustained Contact (Hip/Pelvis) & Saturating EMA & $-1.0 / -3.0$ \\
  Action / Action Rate & Quadratic ($L_2$) & $-0.05 / -0.5$ \\
  Alive Bonus & Constant Indicator & $0.5$ \\
  \bottomrule
\end{tabularx}
\end{table}

\subsection{Motion Tracking Controller}\label{subsec:training}
The motion controller $\pi(a_t | o_t)$ can take one of several forms.
In this work, our aim is to show
that the generated plans are realistic enough that a controller can
use them to effectively navigate the confined environment.
Thus, we present a custom residual RL policy constructed by 
following and adding
to standard approaches~\cite{Liu2025Opt2Skill:Loco-Manipulation}, 
followed by simulation experiments under disturbances in the MuJoCo environment.

Assuming a skill-based control paradigm~\cite{Liu2025Opt2Skill:Loco-Manipulation}, 
we design a learned policy to track
the aforementioned plans for confined spaces. 
We formulate the control task as a Partially Observable Markov Decision Process (POMDP) 
defined by the tuple 
$\mathcal{M} = (\mathcal{S}, \mathcal{O}, \mathcal{A}, \mathcal{P}, \mathcal{R}, \Omega, \gamma)$. 
In an effort to address the sim-to-real gap, we employ an asymmetric actor-critic framework where the 
policy $\pi_\theta(a_t \mid o_t)$ maps noisy local observations $o_t \sim \Omega(\cdot \mid s_t)$ 
to residual joint targets $\delta a_t \in \mathcal{A}$, while the value network 
evaluates based on the full, uncorrupted state $s_t \in \mathcal{S}$.

\textbf{Observations \& Critic State:} The policy observation vector $o_t$ contains tracking errors evaluated relative to the reference guide, including base spatial pose and velocity errors, center-of-mass tracking error, joint position and velocity errors, and relative end-effector tracking errors for hands, knees, and feet relative to the base. Additionally, $o_t$ incorporates noisy proprioception (base velocities, projected gravity, and joint states (see Table~\ref{tab:dr-terms})), previous residual actions, torque history, local obstacle distances, phase progress, and look-ahead windows for the planned contact schedule, forces, surface normals, and end-effector references. The asymmetric critic additionally receives privileged ground-truth contact forces and contact locations during training.


\textbf{Actions \& Rewards:} The policy outputs specify residual joint position targets $q_{\mathrm{target}} = q_{\mathrm{ref}} + \delta a_t$ executed via low-level PD control. The reward terms are listed
in Table~\ref{tab:residual-rewards}. These maximize joint, CoM, and end-effector 
tracking performance, promote contact quality, and penalize residual magnitude and action rates.

\textbf{Domain Randomization \& Terminations:} To maximize robustness, episodes undergo the domain randomization shown in Table~\ref{tab:dr-terms} that includes ground friction, mass scaling, 
and continuous perturbations at the base and hands. 
The mass scaling is applied jointly to the pelvis, torso, and both hip-pitch and knee 
links via a physically consistent pseudo-inertia perturbation 
(mass and inertia scale together, center of mass is unchanged).
Inspired by~\cite{Peng2018DeepMimic:Skills} we randomize the
initial state of the robot to speed up the training.
Episodes are terminated on timeout upon guide exhaustion, root falling, or excessive tilt.

The policy is trained in the Unitree RL mjlab
environment~\cite{robotics_unitree_rl_mjlab_2026} and is
optimized by Proximal Policy
Optimization (PPO). The actor-critic networks are Multi-Layer Perceptrons (MLPs)
with hidden dimensions [1024, 512, 256] and [2048, 1024, 512], respectively.

\section{Results} \label{sec:results}
\begin{figure}
    \centering
    \includegraphics[width=0.8\linewidth]{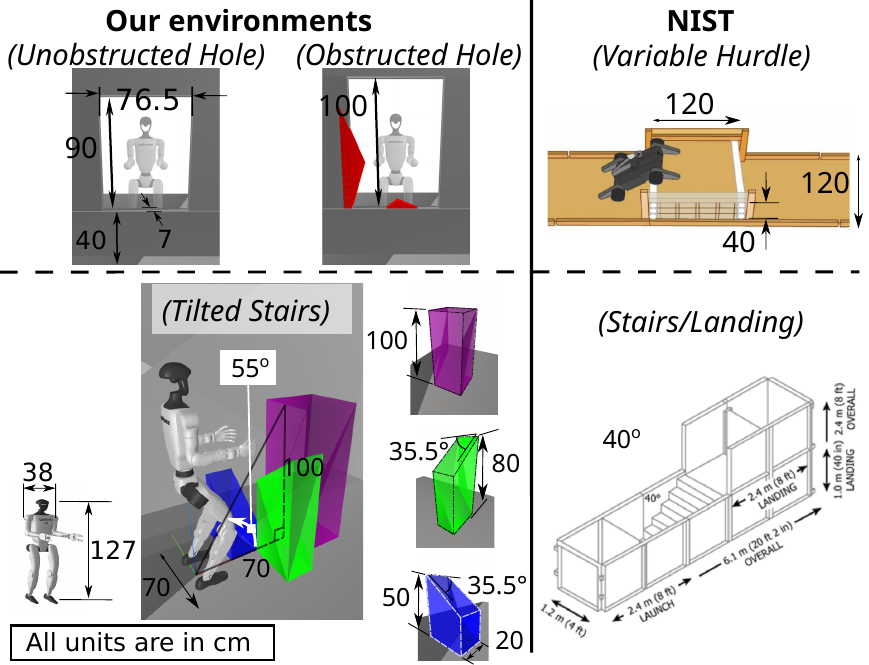}
    \caption{\textbf{Challenging confined environments for benchmarking our WBP.} 
    (Left) Our environments and (right) those proposed by 
    NIST~\cite{2020StandardStairs/Landings, 2021TestObstacles}.
    (Top) The \textit{Unobstructed} and \textit{Obstructed Holes}
    have a step height comparable to the largest hurdle height with a tighter opening width.
    The obstructing obstacles (shown in red) make the hole area non-convex.
    (Bottom) The tilted stairs require the use of palm 
    contacts on the vertical surfaces to balance, we also showcase a steeper staircase slope.}
    \label{fig:experiment-envs}
\end{figure}
%
%


We evaluate the 27-DoF Unitree G1 humanoid across three confined testbeds exceeding NIST 
emergency standards~\cite{2020StandardStairs/Landings, 2021TestObstacles} 
(Fig.~\ref{fig:experiment-envs}): an \textit{Unobstructed Hole}, 
an \textit{Obstructed Hole}, and \textit{Tilted Stairs}. To quantify difficulty, we 
adopt the confinement ratio 
$C_r \triangleq E_{\text{ca}} / A_{\text{ca}}$~\cite{Murphy2014DisasterRobotics}, comparing the void 
cross-sectional area, $E_{\text{ca}}$, to that of the robot, $A_{\text{ca}}$, where $C_r < 2$ 
denotes restricted maneuverability. In its nominal stance, G1 faces $C_r = 1.4$ for the
\textit{Unobstructed Hole}, $C_r = 1.5$ for the \textit{Obstructed Hole}, and $C_r = 2.0$ for 
the \textit{Tilted Stairs}. These provide a lower-bound estimate to terrain
complexity as terrain verticality is not considered in this metric.

The candidate contact sequences consist of non-prehensile palm and foot contacts: 
the \textit{Unobstructed Hole} uses $12\,\text{s}$ to step over the base and $15\,\text{s}$ to 
cross by stepping on the base. The \textit{Obstructed Hole} uses two $15\,\text{s}$ sequences 
stepping on the base, left and right of the obstacle. The \textit{Tilted Stairs} uses 
an $18\,\text{s}$ sequence with opposing hand-foot support. 
To evaluate robustness, we randomly assign different knee modalities and sample 10 
initial base positions uniformly from environment-specific domains around the nominal starting 
positions.

For planning, we assign each contact phase a duration of $3\,\text{s}$,
use hardware torque limits, 
and expand ankle roll limits to $\pm 36^\circ$ for stairs.
All optimizations prioritize feasibility over optimality and run on an 
Intel i7-14650HX CPU with an RTX 5060 GPU. Specifically, our experiments demonstrate: 
(1)~locomotion discovery across confined spaces, 
(2)~failure of standard baselines, 
(3)~necessity of each pipeline stage, and 
(4)~closed-loop execution in physics simulation.

\begin{table*}[t]
\centering
\caption{\textbf{Benchmark Success Rates Across Multiple Confined Spaces.} Comparison of our proposed WBP against baselines and ablations across 10 random initial positions per contact sequence. 
Values for \textbf{Success Rate} \bt{(Mean Solve Time (Std. Dev.) [s])}.
}
\label{tab:master_results}
\begin{tabular}{llccccc}
\toprule
& & \textbf{Tilted Stairs} & \multicolumn{2}{c}{\textbf{Unobstructed Hole}} & \multicolumn{2}{c}{\textbf{Obstructed Hole}} \\
\cmidrule(lr){3-3} \cmidrule(lr){4-5} \cmidrule(lr){6-7}
\textbf{Category} & \textbf{Pipeline Configuration} & Opposing Hand-Foot & Step Over & Step On & Left Step & Right Step \\
\midrule
\rowcolor{gray!15}
\textbf{Proposed} & \textbf{Full Pipeline (Stage 1+2+3)} &  \textbf{10/10} \bt{(174 (57))} & \textbf{10/10} \bt{(125 (31))} & \textbf{10/10} \bt{(180 (41))}  & \textbf{7/10} \bt{(191 (37))}  & \textbf{6/10} \bt{(332 (235))} \\
\midrule
\multirow{2}{*}{\textbf{Baselines}} 
& Spline Avoid Env w/o Knee & 9/10 \bt{(352 (147))} & 0/10 & 0/10 & 0/10 & 0/10 \\
& Lin. Frame Interpolation  & 7/10 \bt{(632 (121))} & 0/10 & 0/10 & 0/10 & 0/10 \\
\midrule
\multirow{2}{*}{\textbf{Ablations}} 
& w/o Stage 2 & \textbf{10/10} 
\bt{(255(79))} & \textbf{10/10} \bt{(131 (20))} & \textbf{10/10} \bt{(189 (37))} & 1/10 \bt{(188)} & 0/10 \\
& w/o dyn. pass in Stage 3  & \textbf{10/10} \bt{(133 (28))} & 7/10 \bt{(140 (44))} & \textbf{10/10} \bt{(163 (102))} & 4/10 \bt{(216 (33))} & 5/10 \bt{(233 (85))} \\
\bottomrule
\end{tabular}
\end{table*}

\subsection{Benchmark Performance in Confined Spaces}
%
%
\begin{figure}[t]
\centering
\includegraphics[width=0.9\linewidth]{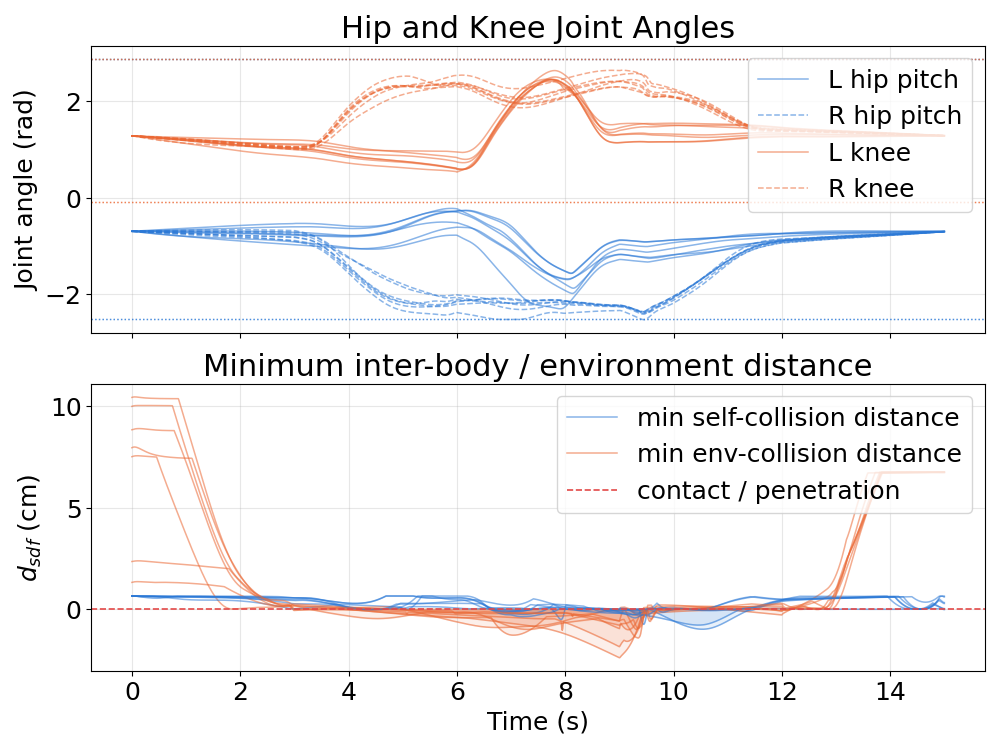}
\caption{ \textbf{Safety margins in planned motions.}
(Top) Joint margins and (bottom) minimum signed distances across all $N=7$ feasible trials for the \textit{Obstructed Hole} (Left Step) scenario. The top- and bottom-most dashed lines denote limits. All generated trajectories exploit kinematically challenging motions while respecting collision (the largest penetration here is of $d_{\mathrm{sdf}} = 2.4~\mathrm{cm}$).}
\label{fig:margins}
\end{figure}
%
%


We show in Table~\ref{tab:master_results} the success rates and 
time it took to solve problem~\eqref{eq:wbp-dt}
with our proposed framework. This evaluation is done
in our benchmark environments, listed in increasing levels
of difficulty: \textit{Titled Stairs} $\rightarrow$ \textit{Obstructed Hole}.
Our approach generates feasible plans for all starting positions
in the \textit{Tilted Stairs} and the \textit{Unobstructed Hole} while
keeping solve times between two and three minutes, on average.
In the \textit{Obstructed Hole}, we find feasible plans
in 65\% of the trials, with compute times averaging 3 and 6 minutes, 
depending on the contact sequence. The high variance in the Right Step sequence is
explained by one trial that lasted 854 seconds to compute. Without it, the
values drop to $\mu=227~\mathrm{s}, ~\sigma=31~\mathrm{s}$.

Not shown in the table are the Stage 2 compute times. 
These kinematic guides take (using the $(\mu, \sigma)$ convention), on average:
$70 (59)$, $78 (9)$, and $128 (61)$ seconds to compute for each environment, respectively.
Representative motions generated by Stages 2 and 3 are provided in the Supplementary 
Material, showing the corrective characteristics that our constrained rigid body planner
provides.

The resulting motions exhibit high and smooth dexterity, despite the
tight open space. Sample motions of the hips and knees traversing the
\textit{Obstructed Hole} are shown in Fig.~\ref{fig:margins}. The top
plot shows the complexity of the task as the robot operates near its joint limits,
while remaining close to its other links and the environment, shown in the bottom plot.
We allow small penetrations (in the range of a couple of centimeters), 
considering that the real-time controller should be able to correct for these
and other uncertainties, as shown in Sec.~\ref{subsec:results_rl}.


\subsection{Baseline Comparisons}
We consider two families of baseline guides in WBP for comparison:
(1) planners that compute kinematically reachable
spline trajectories avoiding environment collisions but excluding knees trajectories,
and (2) planners that account for knee trajectories but perform
a simple linear interpolation between all frames.
The former emulates references generated by planners that rely on
ROMs, i.e., references that are possibly dynamically feasible 
but with low-fidelity in collision avoidance as they disregard knee motions
which implicitly affect foot trajectories.
The latter consists on performing linear interpolation between the contact states:
the simplest references possible when considering additional frames like the knees.

Table~\ref{tab:master_results} indicates that planners disregarding knee
trajectories\footnote{To replicate planners of this form, we remove the knee frames from our
planning frames. This results in spline trajectories for the torso, feet, and
hands, all of which are kinematically reachable.} can lead to
feasible motions in restricted maneuverable spaces such as the \textit{Tilted Stairs}
but largely fail in tighter scenarios such as the \textit{Hole} environments.
This holds for the linear interpolation method at the expense of even higher compute times,
indicating that the optimizer has to work harder to find feasible motions 
further from the poor-quality references. 
Notably, this suggests that when the end-effectors
are within sight from one contact sequence to the next (such as in the
\textit{Tilted Stairs}), the probability of finding a feasible plan is considerably
higher than when the end-effectors are out-of-sight. All of the contact sequences in the
\textit{Hole} environments involve a step over the base. Planning foot trajectories
without guiding the knees in this tight space imposes a tough challenge to the optimizer.
On the other hand, linear interpolation leads to references passing through the base and
colliding with it.

\subsection{Pipeline Ablation Study}

We assess the impact of the different steps of our pipeline. 
Our first ablation skips the volume-aware SCA planning, which is one of our
main contributions. Notably, this ablation is an improved version 
of~\cite{Gonzalez2024GuidingOptimization} due to our collision inflation heuristic
and since our WBP enforces hard inequality
constraints, rather than relying on barrier methods. 
This ablation surpasses most ROM spline-based planners
by including knee trajectories, thus yielding a high success 
rate across the \textit{Tilted Stairs} and the
\textit{Unobstructed Hole}, but finds only one feasible plan in the \textit{Obstructed Hole}.
This shows that our volume-aware SCA guides provide more accurate references in tighter settings.

Our second ablation bypasses the dynamics feasibility pass.
Surprisingly, bypassing it still yields comparable success rates in the
challenging \textit{Obstructed Hole} at the expense of losing some
reliability in the \textit{Unobstructed Hole} stepping over
the base.
This indicates that the dynamics feasibility pass
increases the reliability of the initial guess before performing
a fully constrained WBP.


\subsection{Closed-Loop Controller in Physics Simulation}\label{subsec:results_rl}
We use the feasible plans and their respective domain of initial base positions
to train a residual policy using the structure
detailed in Sec.~\ref{subsec:training}, without modifying it for each environment.
After spawning the robot in its uniformly sampled position, 
the closest plan to its starting position is given as reference for training.
All policies train to convergence within $2\times 10^8$ environment steps (or $2,000$ iterations
when running $4096$ environments) thanks to our high-quality plans.
At this point, our policy achieves $>95\%$ traversal success rate operating under 
full domain randomization and external push perturbations.
The resulting motions are smooth and confidently
traverse the confined spaces, as seen in Fig.~\ref{fig:teaser}.
This training architecture leads to ample coverage while being
able to leverage the different knee styling resulting from our plans, 
as seen in the simulations included in the Supplementary Material.

\section{Conclusion and Future Work}\label{sec:conclusions}
We presented a WBP framework for humanoid locomotion in highly confined environments. 
Rather than relying on traditional point-particle guide abstractions, our core contribution 
is formulating kinematic path planning directly over kinematically reachable rigid volumes. 
By integrating differentiable collision avoidance into a 
reachability-constrained formulation, our framework synthesizes volume-informed guides that 
successfully guide a full-order trajectory optimizer in restricted 
geometries ($C_r < 1.5$) where standard planners fail.
We evaluated multiple candidate contact sequences per environment and 
demonstrated that our planner discovers feasible WBPs that can be further explored in
parallel.
Furthermore, we demonstrated that these plans serve as effective 
references for training a robust residual RL policy, achieving high
traversal success under extensive domain randomization in physics simulation.

Future work includes hardware validation on a physical robot,
using our planner in efficient contact exploration pipelines in search of a general
confined locomotion policy, and extending our strategy to plan 
loco-manipulation with large objects in restricted maneuverable spaces.

\section*{ACKNOWLEDGMENT}
This work was supported by the 
Office of Naval Research (ONR), Award No. N00014-22-1-2204.

\bibliography{references,references_zotero}
\bibliographystyle{ieeetr}

\end{document}